%% file: colm2025_conference.tex
\documentclass{article} 
\usepackage[preprint]{colm2025_conference}

\usepackage{microtype}
\usepackage{hyperref}
\usepackage{url}
\usepackage{booktabs}

\usepackage{lineno}
\usepackage{graphicx}
\usepackage{makecell}
\usepackage{multirow}
\usepackage{wrapfig}
\usepackage{caption}
\usepackage{xspace}
\usepackage{amsmath}
\usepackage{enumitem}              
\usepackage{amssymb} 
\usepackage{titlesec}

\definecolor{darkblue}{rgb}{0, 0, 0.5}
\hypersetup{colorlinks=true, citecolor=darkblue, linkcolor=darkblue, urlcolor=darkblue}

\DeclareRobustCommand{\mybasemethod}{\textsc{ThinkLogit}\xspace}
\DeclareRobustCommand{\mymethod}{\textsc{ThinkLogit-DPO}\xspace}

\titlespacing{\section}{4pt}{3pt plus 2pt minus 1pt}{3pt plus 2pt minus 1pt}

\title{Logit Arithmetic Elicits Long Reasoning Capabilities \\ Without Training}

\author{Yunxiang Zhang\thanks{$\,\,\,$Correspondence to \tt{yunxiang@umich.edu}}\hspace{3pt}\quad Muhammad Khalifa \quad Lechen Zhang\quad Xin Liu \\
\textbf{Ayoung Lee} \quad
\textbf{Xinliang Frederick Zhang} \quad \textbf{Farima Fatahi Bayat}  \quad
\textbf{Lu Wang} \\
Computer Science and Engineering \\
University of Michigan, Ann Arbor
}

\begin{document}

\ifcolmsubmission
\linenumbers
\fi

\maketitle

\begin{abstract}
Large reasoning models (LRMs) can do complex reasoning via long chain-of-thought (CoT) involving cognitive strategies such as backtracking and self-correction. Recent studies suggest that some models inherently possess these long reasoning abilities, which may be unlocked via extra training. Our work first investigates whether we can elicit such behavior without \textit{any} training. To this end, we propose a decoding-time approach, \mybasemethod, which utilizes logits arithmetic \citep{liutuning} to tune a target large LM for long reasoning using a substantially smaller model as guider. We then show that we can further boost performance by training the guider model with preference optimization over correct/incorrect reasoning pairs sampled from both the target and guider model---a setup we refer to as \mymethod. Our experiments demonstrate that \mybasemethod and \mymethod  achieve a relative improvement in pass@1 by 26\% and 29\%, respectively, over four mathematical datasets using the Qwen2.5-32B when guided by R1-Distill-Qwen-1.5B---a model 21x smaller. Lastly, we show that \mybasemethod can transfer long reasoning skills acquired through reinforcement learning, improving pass@1 by 13\% relative compared to the Qwen2.5-32B base model. Our work presents a computationally-efficient method to elicit long reasoning in large models with minimal or no additional training.\footnote{Our code is publicly avaiable at \url{https://github.com/yunx-z/ThinkLogit}.}
\end{abstract}

\section{Introduction}

Large reasoning models (LRMs), such as DeepSeek-R1~\citep{DBLP:journals/corr/abs-2501-12948}, OpenAI o1~\citep{OpenAI2024LearningToReason}, and Qwen3~\citep{QwenTeam2025Qwen3}, have significantly advanced reasoning by leveraging inference-time compute~\citep{DBLP:journals/corr/abs-2408-03314,DBLP:journals/corr/abs-2407-21787}. These models generate very long chain-of-thought (CoT) traces involving planning, reflection, and self-correction~\citep{DBLP:journals/corr/abs-2503-01307}. It is widely believed that such behavior requires training—either through reinforcement learning with verifiable rewards~\citep{DBLP:journals/corr/abs-2501-12948,DBLP:journals/corr/abs-2411-15124,DBLP:journals/corr/abs-2402-03300} or supervised distillation~\citep{DBLP:journals/corr/abs-2501-19393,li2025llms}. However, this training is costly due to the length of reasoning traces and extensive sampling. While such costs are prohibitive for large models, small models can be trained with modest compute~\citep{dang2025reinforcement,deepscaler2025}. This motivates our central question: \textit{\textbf{Can we utilize a small reasoning model to elicit complex reasoning behavior in a larger model at inference time, without training of the larger model?}}

We address this question with a decoding-time technique, enabling a small reasoning model to guide a target model, mainly by manipulating its logits. Specifically, we use logit arithmetic~\citep{liu2021dexperts,liutuning,mitchellemulator,fangiant} to combine the output distributions of both models, allowing the target model to benefit from the guider model’s long-chain-of-thought capabilities, without any additional training. We call this base approach \textbf{\mybasemethod}. 

Furthermore, as the output distribution of both models may substantially differ, we align them by further training the small guider model. Such training uses Direct Preference Optimization~\citep[DPO;][]{DBLP:conf/nips/RafailovSMMEF23} on preference pairs sampled from the guider and target models, making \mybasemethod more \textit{on-policy}, and then apply logit arithmetic using the finetuned guider model. We refer to this approach as \textbf{\mymethod} and show that such training can further boost performance compared to \mybasemethod.

We evaluate our methods on mathematical reasoning problems under two post-training paradigms for small guider LMs: \textbf{supervised fine-tuning (SFT)} and \textbf{reinforcement fine-tuning (RFT)}. In the supervised setting, our results show that using logits from an SFTed model (DeepSeek-R1-Distill-Qwen-1.5B) improves pass@1 of the Qwen2.5-32B base model by an average of 26\% across four mathematical reasoning datasets (\mybasemethod) and by 29\% with the optimized \mymethod guider. Under the RFT scenario, \mybasemethod enhances pass@1 performance by 13\% compared to the Qwen2.5-32B baseline, without any RL training of the 32B model. 
Our work sheds light on the effectiveness of using small reasoning models to guide much larger non-reasoning models, offering a practical and efficient alternative for long CoT reasoning without training.



\section{Related Work}
\label{subsec:related-work}




\paragraph{Long Chain-of-Thought (CoT) Reasoning.}
Large reasoning models, such as OpenAI’s o1 and o3~\citep{OpenAI2024LearningToReason,openai2025o3}, DeepSeek-R1~\citep{DBLP:journals/corr/abs-2501-12948}, and QwQ~\citep{qwq32b}, achieve state-of-the-art results on mathematical and coding benchmarks by generating CoT traces that often extend to thousands of tokens, enabling systematic backtracking, verification, and self-reflection before a final answer is produced \citep{DBLP:journals/corr/abs-2503-01307}. 
One way to elicit such long-form reasoning is through \textbf{reinforcement learning with verifiable rewards}~\citep{DBLP:journals/corr/abs-2411-15124}. Pioneered by Group-Relative Policy Optimization (GRPO) \citep{DBLP:journals/corr/abs-2402-03300} and refined by more stable and token-efficient variants such as DAPO \citep{DBLP:journals/corr/abs-2503-14476} and Dr.~GRPO \citep{liu2025understanding}, this approach optimizes outcome-based rewards for correctness; nevertheless, mounting evidence shows that it mainly re-weights reasoning patterns already latent in the base model \citep{liu2025understanding,yue2025does}.
A complementary line of work demonstrates that the same capability can be acquired with \textbf{data-efficient supervised fine-tuning}. Distilled long CoTs from stronger teacher models allows a student to extend its reasoning length and thus improve accuracy using only about one thousand examples \citep{DBLP:journals/corr/abs-2501-19393,DBLP:journals/corr/abs-2501-11284,DBLP:journals/corr/abs-2502-03387,li2025llms}.
Finally, \textbf{training-free} methods exploit the fact that pretrained LLMs already exhibit long-CoT behaviours~\citep{liu2025understanding,DBLP:journals/corr/abs-2503-01307}.~\cite{tang2025unlocking} inject contrastive long- versus short-CoT representations into hidden states via representation engineering~\citep{DBLP:journals/corr/abs-2310-01405}, whereas~\cite{zhao2025activation} amplify a handful of key neurons at inference. Both techniques, however, require domain-specific long/short traces and white-box access, limiting their applicability in out-of-domain or black-box settings.
\mybasemethod\ sidesteps these constraints entirely. It keeps the target LLM frozen and, at inference time, fuses its logits with those of a lightweight ``guider'' model trained for long reasoning. This logit-fusion strategy recovers long-CoT behavior induced by training-based methods while introducing no additional training cost or handcrafted long-CoT demonstrations.

\paragraph{Decoding Algorithms for LLM Reasoning.}
Decoding-time interventions offer an attractive alternative to full model fine-tuning: they can improve the reasoning capabilities of an off-the-shelf LLM with only a marginal increase in training or inference cost.
The earliest line of work is Chain-of-Thought (CoT) prompting~\citep{DBLP:conf/nips/Wei0SBIXCLZ22}, which simply asks the model to “think aloud.”
Subsequent self-consistency decoding~\citep{DBLP:conf/iclr/0002WSLCNCZ23} samples a set of diverse CoTs and majority-votes over their answers, while later work shows that short CoTs can even be elicited \emph{without} any prompting~\citep{DBLP:conf/nips/0002Z24}.
Crucially, these traces are usually short: they march directly to the answer without back-tracking or verification, and therefore do not unlock the \emph{long-form} reasoning studied in our work.
A second family, guided decoding, biases generation toward correctness using either self-evaluation signals from the model itself~\citep{DBLP:conf/nips/XieKZZKHX23} or an external discriminator~\citep{DBLP:conf/emnlp/KhalifaLLL023}. Accuracy is further improved by best-of-\emph{n} reranking with discriminative reward models that score either the final answer or the reasoning process~\citep{DBLP:journals/corr/abs-2110-14168,DBLP:conf/acl/ZhangKLKLL024,DBLP:conf/iclr/LightmanKBEBLLS24,DBLP:conf/acl/WangLSXDLCWS24}. Generative reward objectives extend this idea and generalize better across tasks~\citep{DBLP:journals/corr/abs-2402-06457,DBLP:conf/iclr/ZhangHBKKA25,wang2025gram,DBLP:journals/corr/abs-2504-16828}. However, all of these methods depend on sampling many complete reasoning traces and scoring them \emph{after} they are generated, which both raises costs and keeps them in the short-CoT regime.
Auxiliary-model approaches modify the output of a frozen \emph{target} model on the fly.
Contrastive decoding subtracts logits from an “amateur’’ model or layer to suppress low-quality outputs~\citep{DBLP:conf/acl/LiHFLEHZL23,DBLP:conf/iclr/ChuangXLKGH24}, while speculative decoding speeds inference by letting a small draft model propose tokens that the expert later accepts or rejects~\citep{DBLP:conf/icml/LeviathanKM23,DBLP:journals/corr/abs-2504-12329,DBLP:journals/corr/abs-2501-19324}.
A closely related strand, \textbf{logits arithmetic}, blends the output distributions of three models token-by-token~\citep{liu2021dexperts,DBLP:conf/emnlp/OrmazabalAA23,DBLP:conf/naacl/ShiHLTZY24}, successfully emulating task-specific fine-tuning~\citep{liutuning,DBLP:conf/nips/FanL0TQCC24}, scaling laws~\citep{mitchellemulator}, unlearning~\citep{DBLP:journals/tmlr/HuangZ0M0PC25} and even overriding safety filters~\citep{DBLP:journals/corr/abs-2401-17256}.
\mybasemethod\ follows this lightweight pathway by using a compact guider model to unlock long-form reasoning in a frozen large model, while \mymethod\ additionally aligns the guider’s distribution with the target model’s, delivering further gains.

\begin{figure*}
  \centering
  \includegraphics[width=0.80\linewidth]{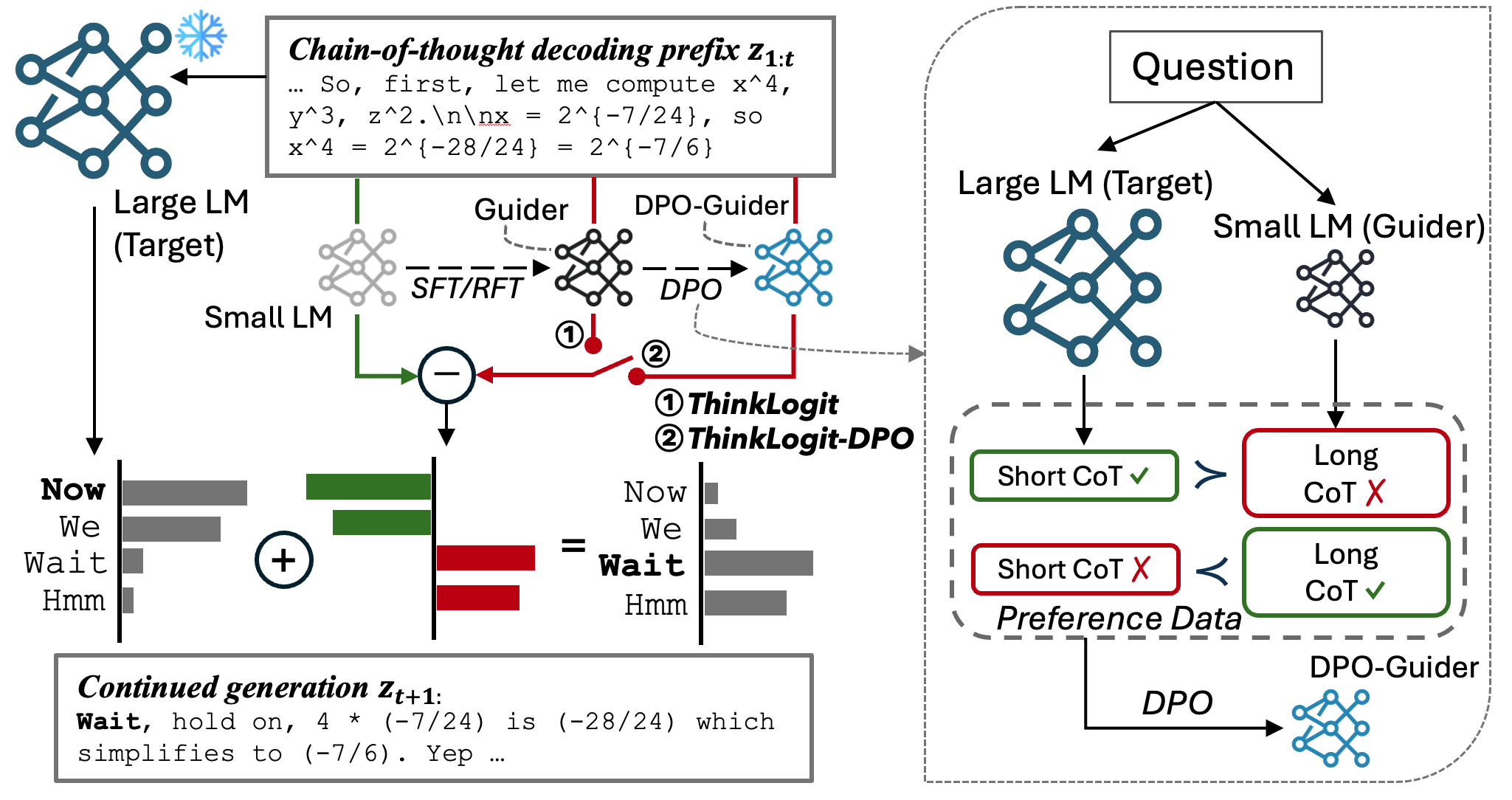}  
  \caption{Overview of our \mybasemethod and \mymethod approaches to elicit long chain-of-thought reasoning from a large pre-trained model.
  }
  \label{fig:overview}
\end{figure*}

\section{Methodology}
\label{sec:method}

Our goal is to elicit long CoT reasoning in a large, frozen language model without expensive training (see Figure~\ref{fig:overview}). We introduce two lightweight decoding-time techniques: \textbf{\mybasemethod}, which transfers long CoT behavior from a small guider via simple logit arithmetic~\citep{liutuning}, and \textbf{\mymethod}, which further refines the guider using Direct Preference Optimization~\citep[DPO;][]{DBLP:conf/nips/RafailovSMMEF23} to align its guidance with the target model.

\paragraph{\mybasemethod.}
Let  
\(
\mathbf{z}_{1:t}=z_1,\dots,z_t
\)  
be the partially decoded sequence of reasoning tokens at step \(t\).  
For any language model \(f\), denote its pre-softmax logits at the next step by
\(
\mathbf{\ell}_{t+1}^{(f)}\! = f(\mathbf{z}_{1:t})
\in\mathbb{R}^{|\mathcal{V}|},
\)
where \(\mathcal{V}\) is the vocabulary.
\noindent We assume three models during inference: \textbf{large (target)} $L$, a pre-trained LLM lacking long CoT capability; \textbf{small base} $S$, a pre-trained model without reasoning fine-tuning; and \textbf{small reasoning (guider)} $S^{\star}$, obtained via SFT or RFT training to $S$.
For a decoding step \(t{+}1\), we compute
$
\mathbf{\tilde\ell}_{t+1}
  = \mathbf{\ell}_{t+1}^{(L)}
  \;+\;
    \alpha \bigl(
        \mathbf{\ell}_{t+1}^{(S^{\star})}
      - \mathbf{\ell}_{t+1}^{(S)}
    \bigr),
\label{eq:logits_arith}
$
where \(\alpha \ge 0\) controls the guidance strength.  
Intuitively, the delta term
\(
\mathbf{\ell}^{(S^{\star})}-\mathbf{\ell}^{(S)}
\)
encodes the probability shift that turns a short-CoT model into a long-CoT one. Intuitively, adding this delta to \(L\) induces analogous long reasoning behavior without altering its weights.\footnote{We empirically observe that applying the delta term at every decoding step can lead to repetitive outputs. To mitigate this, we set \(\alpha = 0\) during an initial warm-up phase before introducing guidance. See Appendix~\ref{subsec:warmup} for implementation details.}

\paragraph{\mymethod.}
The effectiveness of \mybasemethod can be limited by mismatches between the output distributions of the guider and target models.
We therefore construct preference pairs that capture complementary strengths:

\begin{description}[leftmargin=1.4em]
  \item[Type-1:] $(x,\,y^{L\checkmark},\,y^{S\times})$  
        — The \textit{large} model’s correct (short) CoT is preferred over the \textit{small} model’s incorrect (long) one.  
        This encourages the guider to preserve the correctness of the target model and avoid introducing new errors.
  \item[Type-2:] $(x,\,y^{S\checkmark},\,y^{L\times})$  
        — The \textit{small} model’s correct (long) CoT is preferred over the \textit{large} model’s incorrect (short) one, teaching the guider to be more confident at fixing the large model’s reasoning errors.
\end{description}

We gather these pairs from training queries \(x\) by independently sampling CoTs from \(L\) and \(S^{\star}\) and labeling correctness based on the final answer.
Let \(\theta\) denote the parameters of the preference-optimized guider, initialized from \(S^{\star}\).  
We train \(\theta\) with a DPO objective function that mixes the two pair types:
\begin{align}
\mathcal{L}_{\text{DPO}}(\theta)
=\;
&\lambda\,
  \mathbb{E}_{(x,y^{L\checkmark},y^{S\times})\sim\mathcal{D}_1}
  \Bigl[
    \log\sigma\!\bigl(r_\theta(x,y^{L\checkmark}) - r_\theta(x,y^{S\times})\bigr)
  \Bigr]
\nonumber\\[-0.2em]
+\;&(1-\lambda)\,
  \mathbb{E}_{(x,y^{S\checkmark},y^{L\times})\sim\mathcal{D}_2}
  \Bigl[
    \log\sigma\!\bigl(r_\theta(x,y^{S\checkmark}) - r_\theta(x,y^{L\times})\bigr)
  \Bigr],
\label{eq:dpo_two_types}
\end{align}
where  
\(\sigma\) is the sigmoid function,  
$
  r_\theta(x,y)=\beta\!\left[
        \log\pi_\theta (y\mid x)
        -\log\pi_{\text{ref}}(y\mid x)
      \right]
$ is the implicit reward of trajectory \(y\), and  
\(\lambda\in[0,1]\) balances the two datasets \(\mathcal{D}_1\) (Type-1) and \(\mathcal{D}_2\) (Type-2).  
We use \(\lambda{=}\frac{|\mathcal{D}_1|}{|\mathcal{D}_1|+|\mathcal{D}_2|}\) by default, directly concatenating two datasets as DPO training data.
After fine-tuning, we replace \(S^{\star}\) in \mybasemethod with the optimized guider to obtain \mymethod.

\section{Experiments}
\label{sec:exp}

\subsection{Experimental Setup}

\paragraph{Benchmarks.}
We evaluate models on four widely used mathematical reasoning datasets: \textsc{AIME2024}, \textsc{AIME2025}, \textsc{AMC23}, and a subset of 262 challenging problems (levels 4–5) from MATH500~\citep{DBLP:conf/iclr/LightmanKBEBLLS24}, denoted as MATH-hard.  
For MATH-hard, we sample a single completion per query due to its larger size, while for the smaller datasets, we generate 8 independent completions and compute pass@k~\citep{DBLP:journals/corr/abs-2107-03374}. A problem is marked as solved if any of the \(k\) sampled outputs is correct, so pass@k helps reveal a model’s potential to solve a problem. We report both pass@1 and pass@8.

\paragraph{Models.}
Our target model is \textbf{Qwen2.5-32B}~\citep{DBLP:journals/corr/abs-2412-15115}.  
We use two post-trained 1.5B models as guiders. The first one is \textbf{R1-Distill-Qwen-1.5B}~\citep{DBLP:journals/corr/abs-2501-12948}, a version based on Qwen2.5-Math-1.5B~\citep{DBLP:journals/corr/abs-2409-12122} that has been supervised fine-tuned on 800K long-CoT examples distilled from DeepSeek-R1.
The second one is \textbf{One-Shot-RLVR-1.5B}~\citep{wang2025reinforcement}, trained using reinforcement learning with the GRPO algorithm~\citep{DBLP:journals/corr/abs-2402-03300} on Qwen2.5-Math-1.5B~\citep{DBLP:journals/corr/abs-2409-12122} using just two questions.\footnote{\url{https://huggingface.co/ypwang61/One-Shot-RLVR-Qwen2.5-Math-1.5B-pi1_pi13}}

\paragraph{Hyperparameters.}
For all models, decoding is performed with a temperature of 0.6, a maximum length of 8192 tokens, and a guidance strength of $\alpha = 1$.
For preference data collection in \mymethod, we sample 5 completions per model per question from the level 4–5 split of the MATH training set~\citep{DBLP:conf/nips/HendrycksBKABTS21}. We then randomly select 10K preference pairs from the total 50K pairs for DPO fine-tuning. More training details are in Appendix~\ref{subsec:training-details}.

\begin{table}
  \centering
  \small
  \setlength{\tabcolsep}{2pt}  
  \input{main_table}

  \caption{
Comparison of \textbf{pass@1 / pass@8} performance across four mathematical reasoning benchmarks. We omit pass@8 for MATH hard due to budget constraints.  
\textbf{\# T.E.} indicates the number of training examples, and \textbf{\# T.P.} denotes the number of trainable parameters used to elicit long chain-of-thought (CoT) reasoning from the large model.  
Two small reasoning models obtained through supervised fine-tuning (SFT) and reinforcement fine-tuning (RFT) are used as guiders for the target large LM. The best result in each setting without training the large model is marked in \textbf{bold}.  
  }
  \label{tab:reasoning-transfer-results}
\end{table}

\subsection{Main Results and Analysis}
\label{sec:results}

Table\;\ref{tab:reasoning-transfer-results} reports results for both the supervised fine-tuning (SFT) and reinforcement fine-tuning (RFT) settings.
When the 32B target model is augmented with logits from the SFT-trained R1-Distill-Qwen-1.5B guider, \mybasemethod improves the average pass@1 by 26\% relative to the frozen 32B baseline and by 23\% relative to the guider itself, with consistent gains across all four datasets.
Substituting the vanilla guider with the preference-optimized \mymethod{} yields a further boost, reaching a 29\% improvement over the target model.  
Beyond \textit{pass@1}, \mymethod{} consistently increases \textit{pass@8}, confirming that our method \textit{not only increases sample efficiency but also broadens the reasoning boundary}~\citep{DBLP:journals/corr/abs-2504-13837}.

Table\;\ref{tab:reasoning-transfer-results} also lists two strong baselines that fine-tune full parameters: \textsc{s1.1-32B} and \textsc{R1-Distill-Qwen-32B}, which use 1K and 800K distilled examples from DeepSeek-R1 respectively.  
\mymethod{} reaches a pass@8 score of 65.6—just 1.9 points below \textsc{s1.1-32B} (67.5)—while changing only about 0.2\% of the model’s weights.
With 80× less training data, it also cuts 37\% of the pass@1 gap to the strong R1-Distill-Qwen-32B.  
These findings indicate that \textit{lightweight inference-time logit guidance recovers a substantial share of the gains achieved by full-parameter fine-tuning while requiring only a fraction of the data, compute, and parameter updates.}

In the RFT configuration, guidance from One-Shot-RLVR-1.5B~\citep{DBLP:journals/corr/abs-2501-12948} raises the 32B model’s \textit{pass@1} and \textit{pass@8} by 13\% and 23\%, respectively.
Hence, \textit{reasoning skills acquired through data-efficient RL on a compact model propagate to a much larger model at inference time}, entirely avoiding policy updates to the 32B parameters.

\subsection{Ablation Study}
\begin{wraptable}{l}{0.55\textwidth}
  \centering
  \small
  \setlength{\tabcolsep}{2pt}
  \begin{tabular}{l  cc  cc}
    \toprule
    \multirow{2}{*}{Model}
      & \multicolumn{2}{c}{AIME2025}
      & \multicolumn{2}{c}{AMC23} \\
    \cmidrule(lr){2-3} \cmidrule(lr){4-5}
      & Pass@1 & \# Token  
      & Pass@1 & \# Token \\
    \midrule
    Qwen2.5-32B                            
      & 8.3   & 1103 
      & 57.2  & 835 \\
    + budget forcing                      
      & 7.1   & 8003 
      & 49.1  & 7990 \\
    + \mybasemethod        
      & \textbf{19.2} & 5977 
      & \textbf{62.2} & 4343 \\
    \quad w/o warm-up                      
      & 17.9  & 6795 
      & 50.9  & 6442 \\
    \quad $\alpha=0.5$                    
      & 5.8   & 4282 
      & 36.6  & 2399 \\
    \quad $\alpha=1.5$                    
      & 13.3  & 4875 
      & 41.9  & 3381 \\
    \bottomrule
  \end{tabular}
  \caption{Effect of hyperparameter setting in  \mybasemethod on the Pass@1 and average chain-of-thought length (\# Token).
  }
  \label{tab:decoding_ablation}
\end{wraptable}
\paragraph{Ablation on \mybasemethod.}
Table~\ref{tab:decoding_ablation} teases apart the design choices behind \mybasemethod.
Adding a 100-token warm-up before guidance is applied stabilizes early decoding and noticeably boosts accuracy.
The default 
$\alpha=1$ for controlling guidance strength performs best, while deviating from it consistently hurts results.
A budget-forcing~\citep{DBLP:journals/corr/abs-2501-19393} control that merely lengthens outputs by replacing end-of-sentence token with ``Wait'' actually lowers accuracy, confirming that our gains stem from the \emph{quality} of the guidance signal rather than the sheer length of the chain of thought.

\paragraph{Ablation on \mymethod.}
\label{sec:more-ablation}

To further investigate the design decisions in \mymethod{}, we conduct additional ablations on the data construction strategy and training objectives in Table~\ref{tab:decoding_ablation_additional}.

\begin{wraptable}{l}{0.65\textwidth}
  \centering
  \small
  \setlength{\tabcolsep}{2pt}
  \begin{tabular}{l c  cc  cc}
    \toprule
    \multirow{2}{*}{Model}
      & \multicolumn{2}{c}{AIME2025}
      & \multicolumn{2}{c}{AMC23} \\
    \cmidrule(lr){2-3} \cmidrule(lr){4-5}
      & Pass@1 & \# Token  
      & Pass@1 & \# Token \\
    \midrule
    Qwen2.5-32B                            
      & 8.3   & 1103 
      & 57.2  & 835 \\
    + SFT on $y^{S\checkmark}$                      
      & 8.3   & 1535 
      & 61.6   & 1040 \\
    + \mybasemethod        
      & 19.2 & 5977 
      & 62.2 & 4343 \\
    + \mymethod 
      & \textbf{21.7} & 5736 
      & \textbf{63.7} & 4247 \\
    \quad DPO on $(x,\,y^{S\checkmark},\,y^{S\times})$                      
      & 21.2  & 5931 
      & 58.8  & 4371 \\
    + \textsc{ThinkLogit-SFT} \\
    \quad SFT on $y^{S\checkmark}$
      & 20.0  & 5962 
      & 55.6  & 4493 \\
    \quad SFT on $y^{L\checkmark}$
      & 13.3  & 4148 
      & 44.7  & 2872 \\
    \bottomrule
  \end{tabular}
  \caption{Effect of guider's training data and objectives (DPO vs. SFT) in \mymethod, measured by the Pass@1 and average chain-of-thought length (\# Token).
  }
  \label{tab:decoding_ablation_additional}
\end{wraptable}

\textit{Is direct fine-tuning of the target model on distilled long chain-of-thoughts a viable alternative?}
Instead of investing compute in a better guider, one might simply fine-tune the target model on distilled long CoTs from the guider. We therefore fine-tune\footnote{Due to compute constraints for full parameter fine-tuning a 32B model, we use LoRA with a rank size of 256.} Qwen2.5-32B on 10K correct traces from R1-Distill-Qwen-1.5B (``SFT on~$y^{S\checkmark}$'' in Table~\ref{tab:decoding_ablation_additional}). This intervention fails to improve answer accuracy and reasoning length, showing that merely copying CoTs into the large model does not translate into better problem solving, whereas lightweight guidance does.

\textit{Are preference pairs sourced from both the target and the guider necessary to maximize performance?}
To test whether leveraging complementary strengths matters, we construct 10K preference pairs (same amount of training data in \mymethod) using only the guider’s outputs, $(x,\,y^{S\checkmark},\,y^{S\times})$. DPO on this data improves \textsc{AIME2025} to 21.2 but slightly lags behind \mymethod{} (21.7), and it underperforms markedly on \textsc{AMC23} (58.8 vs.\ 63.7). Hence, mixing pairs that highlight \emph{both} the target’s correct answers and the guider’s strong CoTs is crucial for maximal gains.

\textit{Can supervised fine-tuning of the guider equal the effectiveness of preference-based alignment?}
For the purpose of optimizing the guider model, replacing DPO with SFT (\textsc{ThinkLogit-SFT}) on the same number of correct completions (either from guider $y^{S\checkmark}$ or target $y^{L\checkmark}$) lengthens outputs but does not boost accuracy, confirming that preference-driven alignment is far more effective than imitation learning. 

Overall, the best performance arises when the guider is aligned with the target via DPO and trained on mixed preference pairs that capture where each model excels; naive SFT—whether on the target or the guider—fails to match these gains, underscoring the value of preference-driven, complementary data construction in \mymethod{}.

\section{Conclusion and Future Work}
\label{sec:conclusion}


We introduce \mybasemethod{} and \mymethod{}, two decoding-time techniques that unlock long chain-of-thought (CoT) reasoning in frozen LLMs. \mybasemethod{} injects logits from a small, long-CoT guider, boosting accuracy by 26 \% on four math benchmarks for only a 1.1× inference-time parameter cost, while \mymethod{} aligns the guider with target distribution via Direct Preference Optimization for even higher gains. Together they offer a compute-efficient route to deploy long-CoT LLMs. Future work will extend evaluations beyond mathematics, combine heterogeneous model families, and develop context-aware guidance (e.g., adaptive strength~$\alpha$ as in ~\citet{fangiant}) to mitigate the over-thinking problem in long reasoning~\citep{DBLP:journals/corr/abs-2412-21187}.

\section*{Acknowledgments}
This work is supported in part by LG AI Research, Cisco Research, National Science Foundation through grant 2046016, Air Force Office of Scientific Research under grant FA9550-22-1-0099, and computational resources and services provided by Advanced Research Computing (ARC), a division of Information and Technology Services (ITS) at the University of Michigan, Ann Arbor.
We thank the members of the LAUNCH group at the University of Michigan for their discussions and suggestions.

\bibliography{colm2025_conference}
\bibliographystyle{colm2025_conference}

\appendix
\section{Technical Details}
\subsection{Warm-up for Stable Decoding}
\label{subsec:warmup}

We observe that applying logits arithmetic at each decoding step would cause many repetitive generations. To stabilize generations, we defer guidance until a prefix of length \(T_{\text{warmup}}\):
\[
\mathbf{\tilde\ell}_{t+1} =
\begin{cases}
\mathbf{\ell}_{t+1}^{(L)}, & t+1\le T_{\text{warmup}},\\[0.2em]
\mathbf{\ell}_{t+1}^{(L)} + \alpha\bigl(\mathbf{\ell}_{t+1}^{(\text{Guide})}-\mathbf{\ell}_{t+1}^{(S)}\bigr),
& t+1> T_{\text{warmup}},
\end{cases}
\]
where ``Guider'' is the small reasoning model \(S^{\star}\) for \mybasemethod and the DPO fine-tuned model for \mymethod.  
We set \(T_{\text{warmup}}{=}100\) tokens in all experiments.

\subsection{Training Details}
\label{subsec:training-details}

\paragraph{Environment.}
All experiments were conducted using NVIDIA A40
GPUs with 48GB memory. 

\paragraph{LoRA Configuration.}
We applied LoRA~\citep{DBLP:conf/iclr/HuSWALWWC22} for parameter-efficient fine-tuning of the guider model:
\begin{itemize}
  \item Rank: 64
  \item $\alpha_{\text{LoRA}}$: 128
  \item Target modules: \texttt{q\_proj}, \texttt{k\_proj}, \texttt{v\_proj}, \texttt{o\_proj}
  \item Bias: None
\end{itemize}

\paragraph{DPO Training.}
For preference optimization with DPO, we used the following settings:
\begin{itemize}
  \item Batch size: 32 (4 GPUs * 8 Gradient Accumulation)
  \item Epoch: 1
  \item Learning rate: 5e-6
  \item Optimizer: AdamW
  \item Learning rate scheduler: cosine with warmup
  \item Warmup ratio: 0.1
  \item $\beta$ (reward scaling): 0.1
  \item Cutoff length: 8192
  \item Time Cost: 4 hours per run
\end{itemize}

\paragraph{Decoding Setup.}
During inference, we applied the following decoding settings:
\begin{itemize}
  \item Temperature: 0.6
  \item Max tokens: 8192
  \item Top-p: 0.95
  \item Guidance strength $\alpha$: 1.0
  \item Warm-up length: 100 tokens
\end{itemize}


\paragraph{Preference Data Construction.}
We use the level 4–5 subset of the MATH training set~\citep{DBLP:conf/nips/HendrycksBKABTS21} and independently sample 5 completions from both the guider model ($S^{\star}$) and the target model ($L$). Each completion is checked for final-answer correctness against the gold label.\footnote{We extract answers from \texttt{\textbackslash boxed\{\}} and compute exact match with ground-truths based on this script \url{https://github.com/openai/prm800k/blob/main/prm800k/grading/grader.py} by~\cite{DBLP:conf/iclr/LightmanKBEBLLS24}.}

The target model $L$ yields 12,412 correct completions ($y^{L\checkmark}$) and 16,448 incorrect ones ($y^{L\times}$), whereas the guider $S^{\star}$ produces 18,651 correct ($y^{S\checkmark}$) and 10,209 incorrect ($y^{S\times}$) completions.  Forming the Cartesian product for each question gives 11,974 Type-1 preference pairs $\bigl(y^{L\checkmark}, y^{S\times}\bigr)$ and 43,209 Type-2 pairs $\bigl(y^{S\checkmark}, y^{L\times}\bigr)$, for a total of 55,183 pairs used in DPO training.  


\end{document}

%% file: main_table.tex
  \begin{tabular}{lcc|ccccc}
    \toprule
    Model & \makecell[c]{\# T.E.} & \makecell[c]{\# T.P.} & \makecell[c]{AIME\\2024} & \makecell[c]{AIME\\2025} & \makecell[c]{AMC\\23} & \makecell[c]{MATH\\hard} & Average \\
    \midrule
    \textbf{\textit{Large model baselines}}\\
    Qwen2.5-32B (\textbf{\textit{Target}})                                 & - & -     & 14.6 / 40.0 &  8.3 / 26.7 & 57.2 / 90.0 & 50.8 / - & 32.7 / 52.2 \\
    s1.1-32B            & 1K & 32B  & 32.9 / 60.0 & 25.4 / 50.0 & 70.0 / 92.5 & 79.0 / - & 51.8 / 67.5 \\
    R1-Distill-Qwen-32B            & 800K & 32B  & 45.8 / 76.7 & 35.0 / 60.0 & 76.9 / 92.5 & 76.5 / - & 58.6 / 76.4 \\\midrule
    \multicolumn{8}{l}{\textbf{\textit{Supervised Fine-tuned (SFT) LM as Guider}}} \\
    R1-Distill-Qwen-1.5B (\textbf{\textit{Guider}})                        & - & -     & 16.2 / 33.3 & 18.8 / 33.3 & 51.2 / 80.0 & 47.7 / - & 33.5 / 48.9  \\
    Target + \mybasemethod    & 0 & 0      & \textbf{22.5} / 50.0 & 19.2 / 36.7 & 62.2 / \textbf{95.0} & 60.7 / - & 41.2 / 60.6 \\
    Target + \mymethod   & 10K & 78M      & 22.1 / \textbf{60.0} & \textbf{21.7} / \textbf{46.7} & \textbf{63.7} / 90.0 & \textbf{61.1} / - & \textbf{42.2} / \textbf{65.6} \\
    \midrule
    \multicolumn{8}{l}{\textbf{\textit{Reinforcement Fine-tuned (RFT) LM as Guider}}} \\
    One-Shot-RLVR-1.5B (\textbf{\textit{Guider}})     & - & -               & 13.3 / 30.0  & 7.1 / 26.7  & 46.9 / 77.5  & 51.1 / - & 29.6 / 44.7  \\
    Target + \mybasemethod & 0 & 0    & \textbf{17.5} / \textbf{43.3} & \textbf{11.2} / \textbf{36.7} & \textbf{57.2} / \textbf{85.0} & \textbf{61.1} / - & \textbf{36.8} / \textbf{55.0} \\
    \bottomrule
  \end{tabular}